\documentclass[runningheads]{llncs}

\usepackage{preamble}

\begin{document}

\title{The Missing Knowledge Layer in Cognitive Architectures for AI Agents}
\titlerunning{The Missing Knowledge Layer in Cognitive Architectures}

\author{Micha\"el Roynard\orcidID{0000-0001-8469-4131}}
\authorrunning{M. Roynard}

\institute{Independant Researcher\\
  \email{michael.roynard@proton.me}}

\maketitle

\begin{abstract}
  The two most influential cognitive architecture frameworks for AI agents,
  CoALA~\citep{sumers2024coala} and JEPA~\citep{lecun2022jepa}, both lack
  an explicit \Knowledge{} layer with its own persistence semantics. This
  gap produces a category error: systems apply cognitive decay to factual
  claims, or treat facts and experiences with identical update mechanics. We
  survey persistence semantics across existing memory systems and identify
  eight convergence points, from Karpathy's LLM Knowledge
  Base~\citep{karpathy2026llmkb} to the BEAM benchmark's near-zero
  contradiction-resolution scores~\citep{tavakoli2026beam}, all pointing to
  related architectural gaps. We propose a four-layer decomposition
  (\Knowledge, \Memory, \Wisdom, \Intelligence) where each layer has
  fundamentally different persistence semantics: indefinite supersession,
  Ebbinghaus decay, evidence-gated revision, and ephemeral inference
  respectively. Companion implementations in Python and Rust demonstrate the
  architectural separation is feasible. We borrow terminology from cognitive
  science as a useful analogy (the Knowledge/Memory distinction echoes Tulving's
  trichotomy), but our layers are engineering constructs justified by
  persistence-semantics requirements, not by neural architecture. We argue that
  these distinctions demand distinct persistence semantics in engineering
  implementations, and that no current framework or system provides this.

\end{abstract}

\section{Introduction}
\label{sec:intro}

The field of AI agent memory has produced a remarkable proliferation of
systems, from graph-augmented vector
stores~\citep{rasmussen2025graphiti,chhikara2025mem0}, OS-inspired virtual
context managers~\citep{packer2023memgpt}, unified KV+vector+graph
engines~\citep{arqondb2026}, and self-organizing memory
networks~\citep{zhang2025amem}, all attempting to give large
language models persistent state that survives beyond the context window.
We argue that the majority of these systems share a fundamental
\emph{category error}: they conflate knowledge with memory, applying
cognitive decay to factual claims that are not subject to forgetting.

Let us consider NornicDB~\citep{nornicdb2026}, a graph and vector
database targeting AI agents. It implements a three-tier cognitive decay
model: episodic memories receive a 7-day half-life, semantic memories a
69-day half-life, and procedural memories a 693-day half-life. The intent
is biologically inspired. However, a paper's findings do not become less
true after 69~days. A relationship between two concepts does not fade
after a calendar month. What decays is the agent's \emph{attentional
  relevance} to the information (a memory concern, not a knowledge
concern). This system conflates ``I have not accessed this recently''
with ``this is less valuable,'' and those are categorically different
propositions.

Indeed, the conflation is pervasive. Mem0~\citep{chhikara2025mem0}
applies identical CRUD operations to facts and experiences.
Signet~\citep{signet2026} builds a sophisticated entity-aspect-attribute
graph with partial supersession but applies uniform $0.95^{\text{days}}$
decay to all content types. Ori Mnemos~\citep{oriminemos2026} implements
three-zone decay rates but stores all data types in one graph without
formal layer separation. The cost of this conflation is twofold: systems
either \emph{forget what they should remember} (NornicDB applying
storage-level decay to permanent facts) or \emph{remember what they
  should forget} (Hindsight~\citep{latimer2025hindsight} persisting
everything forever without decay).

This category error is not confined to production systems. The two most
influential cognitive architecture frameworks for AI agents both exhibit
it. CoALA~\citep{sumers2024coala} identifies ``semantic memory'' but does
not distinguish its persistence semantics from episodic memory.
JEPA~\citep{lecun2022jepa} has no \Knowledge{} layer at all.

The gap is increasingly visible to practitioners. In April~2026, six
independent community voices articulated the same diagnosis within a
single week on Reddit, spanning three distinct concern tiers. At the
\emph{implementer} tier, the author of a compile-upfront knowledge wiki
acknowledged that ``the compiled layer can drift and start feeling
stale,'' motivating the need for a \Memory{} layer alongside
\Knowledge{}.\footnote{r/Rag, April 2026, u/Astro-Han.} A top-voted
reply to a separate knowledge-compilation thread argued: ``you need to
add a memory layer to your setup, whether it be a bunch of MEMORY.md
files or something fancier. The two work in
tandem.''\footnote{r/Rag, April 2026, u/schneeble\_schnobble (+13
  upvotes).} At the \emph{architect} tier, a practitioner stated the
design principle concisely: ``treat ingestion and interpretation as
probabilistic, but keep storage, state transitions, and supersession
deterministic [\ldots] ontology rules, temporal semantics, and explicit
update policies decide how new information affects existing
knowledge.''\footnote{r/Rag, April 2026, u/JonnyJF.} At the
\emph{governance} tier, an enterprise knowledge architect framed the
problem as: ``what is allowed to compound, what is only a projection,
and what remains the source of record?'' with the invariant that
``projections never silently become the truth they
summarize.''\footnote{r/Rag, April 2026, anonymous OP.} These are
anecdotal signals, not peer-reviewed evidence, but their density and
independence suggest that the architectural gap is recognizable by
practitioners across experience levels.

We analyze
the framework gaps (\cref{sec:background}), propose a four-layer decomposition
with distinct persistence semantics per layer (\cref{sec:four-layers}),
present convergence evidence from 9~independent sources
(\cref{sec:convergence}), briefly describe companion implementations
(\cref{sec:implementations}), and discuss limitations and future work
(\cref{sec:discussion}).

\section{Background}
\label{sec:background}

\subsection{Cognitive Science Foundations}
\label{sec:cogsci}

The four-layer decomposition we propose is an engineering construct, not
a neuroscience model. Nevertheless, it draws on well-established
dissociations in cognitive psychology that motivate treating different
kinds of persistent state differently.
\citet{tulving1972episodic} introduced the canonical
episodic-versus-semantic memory distinction: semantic memory stores
general world knowledge ``not tied to a specific episode,'' while
episodic memory is event-bound with sensory, temporal, and spatial
anchoring. This dissociation is the psychological foundation for our
\Knowledge{}/\Memory{} split.
\citet{cohen1980procedural} demonstrated the procedural-versus-declarative
dissociation through amnesia studies: patients with damaged hippocampi
can learn new motor skills while being unable to form new declarative
memories, establishing that ``knowing how'' and ``knowing that'' are
architecturally distinct. This validates treating \Wisdom{} (procedural,
action-shaping) as a separate layer from \Knowledge{} and \Memory{}
(both declarative).
\citet{bjork1989forgetting} reframed forgetting as an adaptive
retrieval-inhibition mechanism rather than a failure, the position our
\Memory{} layer's Ebbinghaus decay adopts.
\citet{nelson1990metamemory} introduced metamemory (knowledge about
one's own memory processes), which maps onto the \Wisdom{} layer's
self-referential capability.
\citet{mcdaniel2007prospective} synthesized the prospective memory
literature, distinguishing time-based from event-based intention
triggering, both of which our \Memory{} layer addresses.
These references are the same cognitive-science foundations that
\citet{burnell2026deepmind} cite in their recent DeepMind Cognitive
Framework, providing independent anchoring for our architectural
choices.

\subsection{CoALA}
\label{sec:coala}

CoALA~\citep{sumers2024coala} provides the most comprehensive taxonomy of
cognitive capabilities for language agents, decomposing agent cognition
into working memory, episodic memory, semantic memory, and procedural
memory, directly inspired by Tulving's
trichotomy~\citep{hu2025memsurvey,jiang2026anatomymemory}. The framework has become the standard
reference for reasoning about agent memory architecture.

However, CoALA does not distinguish the \emph{persistence semantics} of
semantic memory from episodic memory. Both are classified as ``long-term
memory'' with no formal difference in update mechanism, ownership scope,
or decay behavior. Under CoALA, the fact ``LoRA achieves 95\% of full
fine-tuning quality'' and the experience ``user corrected me about LoRA
yesterday'' inhabit the same architectural category. Yet the first is a
permanent claim that should be superseded (not forgotten) when newer
evidence arrives, while the second is an ephemeral experience that should
decay unless consolidated into a durable behavioral pattern. Simply said,
CoALA's taxonomy correctly \emph{names} the distinction (semantic vs.\
episodic) but does not \emph{operationalize} it with different persistence
mechanics.

\subsection{JEPA}
\label{sec:jepa}

LeCun's ``A Path Towards Autonomous Machine
Intelligence''~\citep{lecun2022jepa} proposes a six-module cognitive
architecture (Perception, World Model, Cost Module, Short-Term Memory,
Actor, and Configurator) centered on the Joint Embedding Predictive
Architecture. The paper is among the most cited position papers in recent
AI research.

JEPA has \textbf{no Knowledge layer}. Factual knowledge about the world
is either (a)~compressed into World Model weights (lossy, unattributable,
and requiring retraining to update) or (b)~held transiently in the
Short-Term Memory buffer (ephemeral, lost on clearance). There is no
persistent, shared, source-attributed factual store. The architecture
provides no mechanism for supersession (recording that a new claim
improves upon an old one while preserving both), no provenance tracking,
and no shared factual store across agents.

Independent critical reviews strengthen this analysis.
\citet{lett2024jepa} identifies that JEPA's configurator is
underspecified, the System~1/System~2 mapping is incorrect, and
Hierarchical JEPA is incompatible with predictive coding.
\citet{bandaru2024jepa} shows that memory is load-bearing for JEPA's
critic: the Cost Module is ``trained from past states and subsequent
intrinsic cost, \emph{retrieved from memory}.'' The critic cannot evaluate
predicted states without access to historical experience, yet JEPA's
Short-Term Memory is an unstructured buffer with no decay policy, no
temporal indexing, and no consolidation mechanism.

It is important to note that JEPA's stated scope is learning architecture
(self-supervised learning via joint embeddings), not agent memory
infrastructure. The gap we identify becomes relevant when JEPA-like
architectures are deployed as cognitive cores for persistent agents: facts
about the world need a home that is neither lossy weight compression nor
ephemeral buffer state. CoALA at least \emph{identifies} semantic memory
as a distinct category, even though it does not separate its persistence
semantics. JEPA does not address persistence at all, which is appropriate
for its scope but insufficient for agent deployment.

\section{The Four-Layer Decomposition}
\label{sec:four-layers}

\begin{figure}[tb]
  \centering
  \includegraphics[width=\columnwidth]{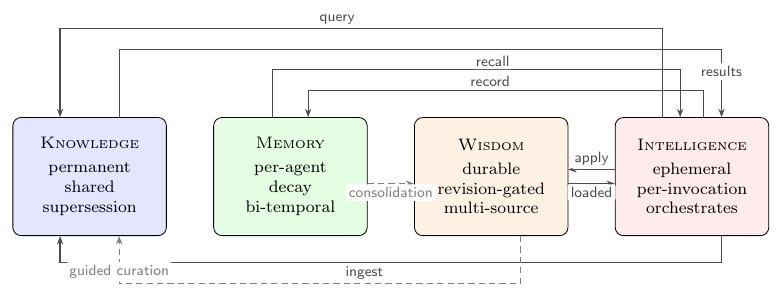}
  \caption{Four-layer cognitive architecture. Solid arrows (top) denote
    working pipelines active during inference. Dashed arrows (bottom) denote
    offline consolidation pipelines. The layers are co-equal substrates with
    distinct persistence semantics, not a strict hierarchy.}
  \label{fig:four-layers}
\end{figure}

We propose that the cognitive substrate of AI agents decomposes into four
layers, each with fundamentally different persistence semantics, update
mechanisms, and ownership scopes (\cref{tab:four-layers},
\cref{fig:four-layers}).

\begin{table}[tb]
  \centering
  \caption{Four-layer decomposition with persistence semantics. Each layer
    requires a different update mechanism. Treating any two identically
    produces the category error described in \cref{sec:intro}.}
  \label{tab:four-layers}
  \small
  \renewcommand{\theadfont}{\bfseries\small}
  \renewcommand{\cellalign}{tl}
  \begin{tabular}{@{} l l l l l @{}}
    \toprule
    \thead{Layer} & \thead{Definition}           & \thead{Persistence} & \thead{Update} & \thead{Scope} \\
    \midrule
    \Knowledge
                  & \makecell{What is                                                                   \\true}
                  & \makecell{Indefinite;                                                               \\supersession}
                  & \makecell{Append-only                                                               \\+ provenance}
                  & Shared                                                                              \\[4pt]
    \hline
    \Memory
                  & \makecell{What                                                                      \\happened}
                  & \makecell{Ebbinghaus                                                                \\decay}
                  & \makecell{Bi-temporal                                                               \\event sourcing}
                  & Per-agent                                                                           \\[4pt]
    \hline
    \Wisdom
                  & \makecell{What                                                                      \\works}
                  & \makecell{Durable;                                                                  \\revision-gated}
                  & \makecell{Evidence-threshold                                                        \\review}
                  & Multi-source                                                                        \\[4pt]
    \hline
    \Intelligence
                  & \makecell{Capacity                                                                  \\to reason}
                  & \makecell{Ephemeral                                                                 \\(inference-time)}
                  & N/A
                  & Per-invocation                                                                      \\
    \bottomrule
  \end{tabular}
\end{table}

\paragraph{Knowledge: ``what is true about the world.''}
Knowledge is factual, structural, and permanent. Facts do not expire;
they get \emph{superseded} by newer evidence, which is a qualitatively
different operation from forgetting. Let us consider a research agent that
has ingested Paper~A: ``LoRA achieves 95\% of full fine-tuning quality at
0.1\% parameters'' and later Paper~B: ``DoRA achieves 97\% at 0.08\%
parameters.'' In a system that conflates knowledge and memory, Paper~A's
claim would decay over time, losing retrievability simply because days
have elapsed. However, Paper~A's finding has not become false. The correct
operation is \emph{supersession}: recording the relationship between the
two claims and marking one as improved upon, while preserving both for
provenance and historical queries. Knowledge is shared across agents (any
agent querying the knowledge base sees the same facts) and carries
provenance: who said it, when, based on what evidence.

\paragraph{Memory: ``what happened, what I was told.''}
Memory is experiential, per-agent, and ephemeral by default. It decays
naturally following a forgetting curve~\citep{yu2026agemem} unless
consolidated. We use Ebbinghaus decay as a simplifying approximation; the
cognitive forgetting literature is considerably richer (interference-based
forgetting, reconsolidation, sleep-dependent consolidation). The
architectural point is that \Memory{} requires \emph{some} decay
mechanism, while \Knowledge{} requires none. One-time observations should not permanently consume
retrieval bandwidth, while recurring patterns should accumulate enough
reinforcement to survive forgetting. Also, memory is context-scoped:
memory about project~A should not bleed into project~B. Every memory
operation produces an immutable event in an append-only event log, with
four timestamps following Graphiti's bi-temporal
model~\citep{rasmussen2025graphiti}: system-created, system-expired,
real-world-valid, and real-world-invalid. This distinction between ``when
did we learn this'' and ``when was this actually true'' is essential for
resolving temporal conflicts.

\paragraph{Wisdom: ``what works, learned from experience.''}
Wisdom consists of pre-compiled behavioral patterns, that is, generalized
lessons extracted from experience. The primary criterion that distinguishes
\Wisdom{} from \Knowledge{} is the \emph{update mechanism}, not the
content type. \Knowledge{} updates via \emph{supersession}: when a new
claim contradicts an old one, both are preserved and linked, with the old
claim marked as superseded. \Wisdom{} updates via \emph{evidence-gated
  revision}: a behavioral directive can only be promoted or modified when
structured evidence (corroboration count, session span, contradiction
absence) crosses a threshold. These are different storage operations that
require different implementations. Content type serves as a secondary
heuristic: verifiable, source-attributed claims default to \Knowledge{},
while behavioral directives derived from experience default to \Wisdom{}.

For the sake of clarity, let us consider four boundary cases.
(1)~``User prefers dark mode'' is \Knowledge{} (a verifiable fact, updated
by supersession if the preference changes).
(2)~``When the user asks about UI, check theme preferences first'' is
\Wisdom{} (a behavioral directive, updated by evidence-gated revision).
(3)~``The user mentioned dark mode yesterday'' is \Memory{} (an ephemeral
observation that should decay unless the pattern recurs).
(4)~``Gradient clipping above 1.0 destabilizes training on ResNet-50'' is
ambiguous: it is both a verifiable empirical finding (\Knowledge) and a
potential behavioral directive (\Wisdom). We resolve this by noting that
the \emph{fact} belongs in \Knowledge{} (with supersession if new evidence
contradicts it), while the \emph{derived directive} (``set gradient
clipping to 1.0 or below'') belongs in \Wisdom{} (with evidence-gated
revision). The same observation can produce entries in both layers with
different persistence semantics, and this is by design: the fact persists
indefinitely, while the directive can be revised when the practitioner's
context changes. An alternative design would collapse \Knowledge{} and
\Wisdom{} into a single layer with a stability-tier field and a
source-attribution flag. This is a viable simplification, but it forces a
single update mechanism to handle both supersession and evidence-gated
revision, which we argue are semantically distinct operations.

Wisdom does not decay:
``never store secrets in git'' does not become less wise over time.
However, wisdom updates via \emph{explicit revision}, not gradual fading.
When a behavioral pattern is superseded (\emph{e.g.}, a user changes their preferred
test framework), the old pattern is retired with provenance and the new
pattern is installed.

Concretely, revision-gating assigns each wisdom entry a stability tier
based on corroborating evidence: entries derived from a single episode are
\emph{predictions} (free to churn), entries corroborated across three or
more independent sessions stabilize as \emph{core} patterns, and entries
that persist without contradiction across ten or more consolidation cycles
earn \emph{anchor} status and resist modification. These thresholds are
configurable and empirically motivated by BaseLayer's finding that 20\% of
facts produces equivalent behavioral fidelity to
100\%~\citep{anthropic2026claudecode}. This tiered model is also motivated
by \citet{cheng2026sycophancy}, who show that RLHF-trained
models affirm user behavior 50\% more than humans, and users rate
sycophantic AI 9--15\% higher even after disclosure. Gating on approval
alone would let sycophantic models promote agreeable-but-incorrect
patterns. Gating on structured evidence prevents this.

\paragraph{Intelligence: ``the capacity to reason, plan, and act.''}
Intelligence is ephemeral. It exists only at inference time and leaves no
direct trace; its effects persist only through the other three layers.
Intelligence orchestrates: it queries knowledge, recalls memory, applies
wisdom, uses tools, and synthesizes a response. Two simultaneous sessions
share no intelligence state.

\paragraph{The organizing principle.}
The key design litmus that emerges from this decomposition is the
distinction between \emph{storage-level} and \emph{query-time} properties.
Recency is a query-time heuristic (the \Intelligence{} layer can boost
recent results when recency matters). Decay is a storage-level mechanism
(the \Memory{} layer applies Ebbinghaus forgetting to experiential facts).
Confusing the two produces systems like NornicDB where knowledge is
subjected to storage-level decay when recency should be a query-time
filter. A system that treats all four layers with identical storage and
retrieval semantics will predictably mishandle at least three of them.

\section{Convergence Evidence}
\label{sec:convergence}

The four-layer decomposition is not merely a theoretical proposal.
Multiple independent sources, spanning academic benchmarks, production
systems, industrial-lab reports, and practitioner projects, converge on
related architectural gaps without coordinating
(\cref{tab:convergence}). We note that these sources share intellectual
heritage (several trace to Tulving's trichotomy), so the convergence is
downstream of common ancestors rather than fully independent.
Nevertheless, the pattern is suggestive: each source independently
discovers that flat persistence semantics are insufficient, even if none
proposes the same four-layer resolution we do.

\begin{table*}[tb!]
  \centering
  \caption{Convergence points. Each source independently discovers or
    validates a component of the four-layer decomposition. The ``What's
    missing'' column shows which layers or persistence semantics the source
    still lacks.}
  \label{tab:convergence}
  \small
  \begin{tabularx}{\textwidth}{@{}l X X@{}}
    \toprule
    \textbf{Source} & \textbf{What they found}                                                                                          & \textbf{What's missing} \\
    \midrule
    \multicolumn{3}{@{}l}{\emph{Published / peer-reviewed}}                                                                                                       \\
    \midrule
    \makecell[tl]{DeepMind Cognitive                                                                                                                              \\Framework~\citep{burnell2026deepmind}}
                    & 10-faculty taxonomy; Working Memory under Executive Functions, not Memory; memory/learning distinction
                    & Taxonomy only; no persistence semantics, no implementation                                                                                  \\
    \hline
    Hindsight~\citep{latimer2025hindsight}
                    & 5-level data hierarchy (entities $\to$ facts $\to$ observations $\to$ mental models), BEAM SOTA 64.1\%/10M tokens
                    & No forgetting, no decay, no bi-temporal, no supersession                                                                                    \\
    \hline
    BEAM~\citep{tavakoli2026beam}
                    & Contradiction resolution ${<}0.05$, temporal reasoning $0.12$ across \emph{all} systems at 10M tokens
                    & Reveals gap, does not propose architectural solution                                                                                        \\
    \hline
    \makecell[tl]{Gulli \& Sauco                                                                                                                                  \\~\citep{gulli2025agenticai}}
                    & Google CTO reference (424p, Springer): two-tier memory (context + vector store)
                    & No temporal model, no forgetting, no \Wisdom{}                                                                                              \\
    \midrule
    \multicolumn{3}{@{}l}{\emph{Industry / production systems}}                                                                                                   \\
    \midrule
    Karpathy LLM KB~\citep{karpathy2026llmkb}
                    & \Knowledge{} layer: compilation, querying, maintenance loops (April 2026)
                    & No \Memory, no \Wisdom, no temporal semantics                                                                                               \\
    \hline
    Claude Code~\citep{anthropic2026claudecode}
                    & 4-type taxonomy (user/feedback/project/reference), 4-stage consolidation pipeline
                    & No supersession, no temporal validity, no structured facts                                                                                  \\
    \hline
    Mengram~\citep{mengram2026}
                    & Independent Tulving trichotomy: separate pipelines for semantic/episodic/procedural
                    & No persistence semantics per type                                                                                                           \\
    \hline
    Mastra OM~\citep{mastra2026om}
                    & LongMemEval SOTA (94.87\%); multi-session ceiling at 87.2\% across all systems
                    & No cross-session persistence, no supersession, no layer separation                                                                          \\
    \hline
    Google AOMA~\citep{google2026aoma}
                    & LLM-driven 30-min consolidation loops generating meta-insights from raw memories
                    & No temporal model, no forgetting, no layer separation, O($n$) retrieval                                                                     \\
    \bottomrule
  \end{tabularx}
\end{table*}

The pattern across these convergence points is consistent: teams and
practitioners independently discover the \emph{need} for typed cognitive
data with different persistence semantics, but no existing system provides
the full decomposition. Karpathy builds a \Knowledge{} layer while missing
the other three. Hindsight builds the most sophisticated retrieval system
(BEAM SOTA at 64.1\%) yet persists everything forever with no decay and no
supersession. Claude Code's 4-type taxonomy partially distinguishes
\Knowledge{} from \Memory{} but stores both as markdown blobs with
identical persistence semantics. Also, two recent systems independently
validate consolidation as a first-class operation: Mastra's Observational
Memory~\citep{mastra2026om} achieves LongMemEval SOTA via compression
rather than retrieval, but hits a multi-session ceiling at 87.2\% because
observations vanish when the session ends (no persistent \Memory{} layer).
Google's Always-On Memory Agent~\citep{google2026aoma} runs 30-minute LLM
consolidation loops that generate meta-insights from raw memories,
independently converging on the DreamCycle concept we formalize, but
without temporal modeling, forgetting, or layer separation.

Furthermore, practitioner communities have independently arrived at
similar distinctions. On the r/AIMemory forum, one user unpromptedly
distinguished wisdom from memory: ``It sounds like you're describing
`wisdom' which is another component alongside
memory.''\footnote{r/AIMemory, ``Memory as a Harness: Turning Execution
  Into Learning,'' March 2026, user avwgtiguy.} Another observed that
existing benchmarks ``test whether a system can find or apply what was
said,'' not ``whether it actually built coherent
knowledge.''\footnote{r/AIMemory, benchmark discussion thread,
  March 2026, user PenfieldLabs. See also their LoCoMo audit:
  \url{https://github.com/dial481/locomo-audit}.} A third proposed that
``memory retrieval should be model-agnostic, and the harness layer handles
formatting/routing,'' independently validating the consumer trait
abstraction.\footnote{r/AIMemory, same thread as footnote 1, user
  Time-Dot-1808.}
These are anecdotal signals, not peer-reviewed evidence, but they suggest
that the gap is recognizable by practitioners without exposure to our
framework.

The BEAM benchmark~\citep{tavakoli2026beam} provides the strongest
empirical evidence. At 10M~tokens (where context stuffing physically
cannot work), the abilities where \emph{all} systems score worst are
contradiction resolution ($<$0.05), temporal reasoning (0.12), and
knowledge update (0.26--0.39). These are precisely the abilities that
require \emph{architectural} solutions (supersession, bi-temporal
modeling) rather than better retrieval. This pattern suggests that
persistence semantics, not retrieval quality, is the primary bottleneck.

\paragraph{External validation from DeepMind.}
The strongest external validation comes from DeepMind's Cognitive
Framework~\citep{burnell2026deepmind}, a 32-page technical report
proposing a 10-faculty cognitive taxonomy for measuring progress toward
AGI. Their \S7.5~Memory subdivides into Semantic, Episodic, Procedural,
Prospective, and Forgetting sub-faculties. Crucially, they place Working
Memory under \S7.8~Executive Functions, \emph{not} under Memory, on the
explicit grounds that ``working memory involves the coordination of
multiple faculties including memory, attention, and sometimes reasoning.''
This independently mirrors our \Intelligence{}-versus-\Memory{} split:
DeepMind recognize that the colloquial term ``memory'' conflates runtime
context with durable persistence, and structurally separate them. Their
\S7.5 opening paragraph states that ``learning is focused on the
acquisition of new knowledge, whereas memory is concerned with the
ability to maintain that knowledge over time [\ldots] a failure to
update semantic knowledge despite being able to successfully recall
already stored knowledge would be considered a failure of learning,
while forgetting information over time that was initially successfully
learned would be a failure of memory.'' This is the category error
framed in the same terms as \cref{sec:intro}, stated by a first-party
industrial lab. The 10-faculty taxonomy compresses onto our four layers
via shared persistence semantics: Perception, Generation, Attention,
Working Memory, Reasoning, and Problem-Solving collapse to
\Intelligence{} (all ephemeral runtime); Semantic Memory maps to
\Knowledge{}; Episodic, Prospective, and Forgetting map to \Memory{};
Procedural Memory, Metacognition, and Executive Functions map to
\Wisdom{}. This compression is a simplification, not an exact mapping:
Attention and Working Memory have distinct properties from Reasoning in
DeepMind's framework, and collapsing six faculties into \Intelligence{}
loses granularity that may matter for fine-grained capability evaluation. The ``jagged profile'' diagnostic~\citep{morris2026jaggedness}
complements this: monolithic agent architectures produce uneven cognitive
capability distributions precisely because they conflate layers with
fundamentally different persistence semantics.

\paragraph{Parallel community convergence (April 2026).}
Beyond the practitioner quotes cited above, the April 2026 landscape
surfaced several independent systems converging on components of the
four-layer thesis without coordinating. rohitg00's LLM Wiki
v2~\citep{rohitg002026llmwiki}, a gist forking Karpathy's pattern,
independently proposes confidence scoring with time-decay, explicit
supersession, type-appropriate Ebbinghaus decay rates, and a four-tier
consolidation pipeline, making it the closest informal statement of the
thesis. Hermes Agent~\citep{nous2026hermesagent} (Nous Research) ships
the closest existing implementation of the \Wisdom{}-layer consolidation
loop: procedural skill documents written back from task execution in
real time following the \texttt{agentskills.io} open standard, though it
lacks a bi-temporal substrate.
Semantica~\citep{hawksight2026semantica} and
MinnsDB~\citep{minns2026minnsdb} are the closest architectural cousins:
both implement bi-temporal knowledge graphs (Semantica with Allen
Interval Algebra for deterministic temporal consistency, MinnsDB with
\texttt{WHEN}/\texttt{AS OF} query clauses), though neither separates
persistence semantics per cognitive type.
Frona~\citep{fronalabs2026frona} is the first Rust-language peer,
validating the choice of Rust for memory-layer engines.
Papr's schema-policy DSL~\citep{papr2026sdk} provides
vocabulary directly applicable to the \Wisdom{} revision gate:
declarative promotion predicates that decouple graph structure from
resolution behavior. None of these systems provides the full four-layer
decomposition; each addresses one or two components. The density of
convergent work in a single month suggests that the architectural
pattern is emerging independently across the community.

\paragraph{Why four layers?}
A natural question is whether the decomposition could use fewer or more
layers. A three-layer model (dropping \Wisdom) would merge behavioral
directives with factual knowledge, but these require fundamentally
different update mechanics: facts supersede via evidence, while directives
require evidence-gated revision with stability tiers. To illustrate the
failure mode concretely: in a merged K+W store, a sycophantic pattern
(``always agree with the user's architectural preferences'') could be
promoted to anchor status because it is verifiable (the user did
consistently prefer this) and source-attributed, bypassing the
evidence-gated review that would catch it in a separate \Wisdom{} layer.
Conversely, a five-layer model (splitting \Knowledge{} into factual and
relational substrates) would add complexity without a corresponding
difference in persistence semantics, as both subfactors still require
supersession and provenance. The four-layer count is not a priori
necessary. It is the minimum decomposition where each layer requires a
\emph{different} persistence mechanism: supersession, decay,
revision-gating, and ephemerality respectively. An analogy from human
cognition illustrates why merging \Knowledge{} and \Wisdom{} specifically
is problematic: a student taking an exam uses consolidated skills
(\Wisdom{}, such as how to integrate or solve equations, which do not
decay), factual knowledge (\Knowledge{}, such as ``Paris is the capital of
France,'' which supersedes if a capital moves but does not fade), episodic
memories (\Memory{}, such as what a teacher said last week, which decays
unless rehearsed), and inference-time reasoning (\Intelligence{}, the exam
itself, which is ephemeral). Collapsing skills and facts into one layer
would require a single update mechanism to handle both ``$2+2=4$''
(permanent, not subject to revision) and ``prefer integration by parts for
this class of problems'' (durable but revisable when better techniques are
learned). These are categorically different persistence requirements.

We note that grey literature (blog posts, tweets, GitHub repositories)
constitutes a significant fraction of our references. This is an inherent
property of the field's pace: agent memory systems are evolving on a
weekly cadence, and many of the most relevant contributions exist only as
open-source repositories or practitioner posts. We cite peer-reviewed work
where available and flag non-peer-reviewed sources explicitly.

\section{Companion Implementations}
\label{sec:implementations}

In order to demonstrate that the four-layer architectural separation is
feasible, we contribute two companion implementations that realize the
\Knowledge{} and \Memory{} layers respectively.\footnote{The \Wisdom{}
  layer materializes across multiple substrates: model weights (frozen
  wisdom from training), configuration files (user-curated rules loaded at
  session start), and promoted behavioral directives (a key-value store
  with stability-tier metadata and a revision log). \Intelligence{} is the
  model itself.}

\paragraph{knowledge-base} (\KB)\footnote{\url{https://github.com/dutiona/knowledge-base}}: a Python MCP server backed by SQLite
with \texttt{sqlite-vec} for vector search and FTS5 for full-text
retrieval. It exposes 46 MCP tools covering the full lifecycle (ingestion,
structure extraction, entity and relationship management, hybrid search
with RRF fusion and stage-2 reranking, and conclusion tracking with
supersession). Facts carry provenance metadata and support supersession: old
claims are never deleted but linked to their successors. The system passes
338+ tests.

\paragraph{memory-engine} (\ME)\footnote{\url{https://github.com/dutiona/memory-engine}}: a Rust crate using SQLite for
persistence, in-process HNSW for vector search, and Petgraph for graph
traversal. It implements Ebbinghaus forgetting with configurable
half-lives, bi-temporal fact storage (four timestamps per fact), scoped
contexts for project isolation, and five consumer traits
(\texttt{EmbeddingProvider}, \texttt{SummaryGenerator},
\texttt{ConflictArbiter}, \texttt{PersistenceClassifier},
\texttt{Reranker}) that carry zero network or LLM dependencies, as all
intelligence is consumer-provided. The engine also ships first-class
prospective memory primitives (\texttt{list\_due},
\texttt{next\_due\_time}, \texttt{surfaced\_at}) implementing time-based
intention triggering~\citep{mcdaniel2007prospective}, mapping directly
onto DeepMind's \S7.5.4 Prospective Memory~\citep{burnell2026deepmind}.
The system passes 486~tests.

The architectural separation \emph{is} the contribution, not the search
quality of either system. These implementations demonstrate that
\Knowledge{} and \Memory{} can be realized as separate systems with
different persistence semantics, different update mechanics, and different
ownership models, and that doing so is practical at the library level
without requiring heavy infrastructure.

A notable architectural invariant distinguishes both implementations from
every system surveyed in \cref{sec:convergence}: all core operations
(decay computation, graph traversal, retrieval fusion, supersession
bookkeeping, consolidation scheduling) use deterministic algorithms with
zero LLM calls and zero network dependencies. LLM operations (entity
extraction, fact validation, wisdom promotion) happen exclusively at the
consumer layer, invoked by the calling agent with its own provider.
Indeed, every competing system that performs consolidation or structured
extraction requires LLM calls \emph{inside} the memory system itself
(Mastra's Observer/Reflector agents, Google AOMA's Gemini-based
consolidation, claude-mem's separate Claude session). Our LLM-free engine
guarantees predictable latency, bounded memory footprint ($<$5\,MB RAM),
and offline operation.

\paragraph{Pilot evaluation.}\footnote{Experiment code and data: \url{https://github.com/dutiona/papers-material}.}
In order to test whether the architectural separation changes observable
outcomes, we ran a focused pilot on the BEAM benchmark's 100K-token
split~\citep{tavakoli2026beam}. We selected the two ability categories
where all systems score worst (contradiction resolution and temporal
reasoning) and compared two conditions: \emph{typed routing}, where an
oracle classifier directs queries to \KB{} (supersession-aware) or \ME{}
(bi-temporal) based on the ground-truth BEAM category, and a \emph{flat
  baseline}, where the same queries hit a single undifferentiated
FTS5 store with no type distinction. Both conditions used the same
local model (Gemma~4 26B) for generation and scoring, with 80~questions
from 20~conversations per condition (\cref{tab:pilot}).

\begin{table}[tb!]
  \centering
  \caption{BEAM pilot results (100K split, $N=80$ per condition). Typed
    routing to \KB{} or \ME{} based on query type vs.\ a flat baseline.}
  \label{tab:pilot}
  \small
  \begin{tabular}{@{} l c c c @{}}
    \toprule
    \textbf{Category}        & \textbf{Typed} & \textbf{Flat}  & \textbf{$\Delta$} \\
    \midrule
    Contradiction resolution & 0.500          & 0.394          & +0.106            \\
    Temporal reasoning       & 0.425          & 0.275          & +0.150            \\
    \midrule
    \textbf{Overall}         & \textbf{0.463} & \textbf{0.334} & \textbf{+0.128}   \\
    \bottomrule
  \end{tabular}
\end{table}

Typed routing improves overall accuracy by +0.128 (46.3\% vs.\ 33.4\%,
bootstrap 95\% CI on $\Delta$: $[0.04, 0.22]$, McNemar $p = 0.035$).
The largest gain is on temporal reasoning (+0.150), where the
memory-engine's bi-temporal filtering surfaces chronologically ordered
facts that the flat store returns unordered. Contradiction resolution
gains +0.106, as the knowledge-base's supersession-aware retrieval filters
stale claims. Also, a two-conversation comparison with a heuristic
keyword-based router (instead of the oracle) reverses the typed advantage
($\Delta=-0.125$), confirming that routing accuracy is load-bearing:
architectural separation helps only when queries reach the correct store.

The pilot has clear limitations: small sample ($N=80$), two categories
only, no ablation separating routing from store semantics, FTS-only
retrieval (no vector search), and a local 26B model for both generation
and scoring. Nevertheless, the result is directionally consistent with the
paper's thesis: the abilities where flat stores fail worst are precisely
those where typed persistence semantics provide measurable improvement.

\section{Discussion and Limitations}
\label{sec:discussion}

\paragraph{Why this matters.}
Every system that persists information for AI agents must choose how to
handle persistence semantics. The choice is often implicit and uniform
(identical CRUD for all data types) but it is still a choice, and it
produces predictable failures. The four-layer decomposition makes this
choice explicit: facts get supersession, experiences get decay, behavioral
patterns get evidence-gated revision, and reasoning is ephemeral. Making
the choice explicit does not solve every problem in agent memory, but it
prevents the category errors that currently plague the field.

\paragraph{The null hypothesis.}
The strongest counter-argument to structured external memory comes from
the practitioner community: ``All you are doing is memory with extra steps
and burning huge blocks of context to read all that memory. AI's don't
improve, no matter what you put on disk.'' This objection has two
components. First, the empirical claim that external memory does not
improve performance is falsified by existing benchmarks: Mastra's
Observational Memory~\citep{mastra2026om} reports 94.87\% on LongMemEval
where the baseline scores 33\% (a near-3$\times$ improvement). However,
a recent community audit~\citep{mempalace2026drama} has revealed serious
methodological issues: LoCoMo's answer key is 6.4\% factually incorrect,
its LLM judge accepts 63\% of intentionally wrong answers, and
LongMemEval's per-question context fits in a single window, allowing
systems to bypass retrieval entirely. These findings render all prior
LongMemEval and LoCoMo scores provisional. Nevertheless, the directional
result stands: structured memory outperforms raw context.
Mastra is a flat memory store, which raises a follow-up question: if a
flat store achieves strong retrieval scores, why do we need layer
separation? The answer is that these benchmarks
test retrieval accuracy, not persistence correctness. A flat store can
retrieve the right fact today, but it cannot handle the case where a fact
was superseded yesterday and the old version should still be queryable for
provenance. BEAM's contradiction-resolution scores ($<$0.05 across all
systems) measure precisely this capability, and there flat stores fail.

Second, and more importantly, the ``extra steps'' objection is partially
correct, but it is an argument \emph{for} our architecture, not against
it. Indeed, a flat, undifferentiated memory store where facts, preferences,
and ephemeral observations use identical semantics \emph{is} noise with
extra steps. Thanks to layer-separated persistence semantics, each
retrieval targets the correct substrate: knowledge queries return
source-attributed facts, memory queries return decay-aware experiences, and
wisdom is pre-loaded at session start without retrieval cost at all.

\paragraph{Relationship to broader frameworks.}
\cref{tab:gap} positions the four-layer decomposition against CoALA and
JEPA. The key column is ``persistence distinction'': CoALA names the
Knowledge/Memory boundary but does not operationalize it, while JEPA does
not identify it as a concern.

\begin{table}[tb!]
  \centering
  \caption{Gap analysis: how CoALA and JEPA handle each cognitive substrate
    versus the four-layer decomposition.}
  \label{tab:gap}
  \small
  \renewcommand{\cellalign}{tl}
  \begin{tabular}{@{} l l l l @{}}
    \toprule
     & \textbf{CoALA}               & \textbf{JEPA} & \textbf{Four-Layer} \\
    \midrule
    \Knowledge
     & \makecell{Semantic mem.                                            \\(no distinct\\persistence)}
     & \makecell{In weights (lossy)                                       \\or buffer\\(ephemeral)}
     & \makecell{Supersession                                             \\+ provenance} \\[4pt]
    \hline
    \Memory
     & \makecell{Episodic mem.                                            \\(same persistence\\as semantic)}
     & \makecell{Short-Term Mem.                                          \\(unstructured\\buffer)}
     & \makecell{Ebbinghaus                                               \\+ bi-temporal} \\[4pt]
    \hline
    \Wisdom
     & Procedural mem.
     & \makecell{World Model                                              \\weights}
     & \makecell{Revision-gated                                           \\+ evidence} \\[4pt]
    \hline
    \Intelligence
     & Working mem.
     & \makecell{Configurator                                             \\+ Actor}
     & \makecell{Ephemeral                                                \\orchestration} \\
    \midrule
    \makecell[l]{\textbf{Persistence}                                     \\\textbf{distinction}}
     & \makecell{Named but not                                            \\operationalized}
     & Not identified
     & \makecell{\textbf{Explicit}                                        \\\textbf{per layer}} \\
    \bottomrule
  \end{tabular}
\end{table}

\paragraph{Limitations.}
Despite the theoretical grounding and convergence evidence, this design
comes with some limitations. First, this paper is a position paper: the
companion implementations demonstrate feasibility, not superiority. We do
not present large-scale empirical evaluation on standardized benchmarks,
and that is deferred to future work. Second, the convergence evidence
includes practitioner community signals (footnotes in
\cref{sec:convergence}) that are not peer-reviewed. We include them as
evidence of independent discovery, not as authority. Third, the \Knowledge/\Wisdom{} boundary remains a
design choice: the same observation can produce entries in both layers
(see the gradient-clipping example in \cref{sec:four-layers}), and a
single-layer alternative with stability tiers is a viable simplification.
Fourth, the four-layer decomposition is a design framework, not a formal
model with provable properties. Its value lies in preventing category
errors, not in mathematical guarantees.

\paragraph{Future work.}
Three directions follow directly. First, scaling the pilot evaluation
(\cref{tab:pilot}) to the full BEAM benchmark at 1M and 10M~tokens with
a learned router (replacing the oracle classifier), testing whether
architectural separation improves contradiction resolution and temporal
reasoning scores at scale. Second, a memory-architecture-specific
benchmark (MemArch-Bench) that tests supersession correctness,
bi-temporal point-in-time accuracy, type-appropriate decay, and
context-poisoning resistance, that is, architectural properties that no
existing benchmark evaluates~\citep{tavakoli2026beam}. The recent
benchmark governance crisis~\citep{mempalace2026drama}, which
invalidated direct comparability of LongMemEval and LoCoMo scores across
systems, makes this benchmark an urgent infrastructure need for the
field. Third, a closed-loop memory-to-wisdom consolidation pipeline
(DreamCycle) that promotes recurring patterns from \Memory{} to
\Wisdom{} with full provenance tracking.

\section{Conclusion}
\label{sec:conclusion}

We have identified a missing \Knowledge{} layer in the two most
influential cognitive architecture frameworks for AI agents. CoALA names
the Knowledge/Memory distinction but does not operationalize it with
different persistence semantics. JEPA does not identify it as a concern at
all. This gap produces a category error that is pervasive across the
field: systems apply cognitive decay to facts, or treat facts and
experiences with identical update mechanics. A four-layer decomposition
(\Knowledge, \Memory, \Wisdom, \Intelligence) with distinct persistence
semantics per layer resolves this error. Nine convergence points, from DeepMind's Cognitive
Framework to BEAM's near-zero contradiction-resolution scores, validate
that the gap is widely recognized across industrial-lab reports,
published benchmarks, production systems, and practitioner projects. The
architectural separation is feasible: companion implementations in Python
(338+ tests) and Rust (486~tests) demonstrate it at the library level
without heavy infrastructure.

\clearpage

\bibliographystyle{splncs04}
\bibliography{../shared/references}

\begin{thebibliography}{10}
\providecommand{\url}[1]{\texttt{#1}}
\providecommand{\urlprefix}{URL }
\providecommand{\doi}[1]{https://doi.org/#1}

\bibitem{arqondb2026}
{AlbericByte}: {ArqonDB}: Unified {KV}+vector+graph engine for agent memory.
  GitHub (2026), \url{https://github.com/AlbericByte/ArqonDB}, rust. Causal DAG
  with temporal edges (uni-temporal validity intervals), state branching,
  custom LSM-tree + HNSW. Still conflates all cognitive layers into a single
  ``agent state'' abstraction.

\bibitem{mengram2026}
{alibaizhanov}: Mengram: Three-type memory extraction for {LLMs}. GitHub
  (2026), independent convergence on Tulving's trichotomy
  (semantic/episodic/procedural) with separate extraction pipelines per type.

\bibitem{anthropic2026claudecode}
{Anthropic}: Claude code memory architecture. Claude Code v2.1.59+ (2026), six
  memory subsystems with 4-type taxonomy (user/feedback/project/reference). No
  supersession, no temporal validity, no structured facts. Most widely deployed
  AI memory system.

\bibitem{bandaru2024jepa}
Bandaru, R.: Deep dive into {Yann LeCun's} {JEPA}. GitHub Pages blog (2024),
  \url{https://rohitbandaru.github.io/blog/2024/07/01/deep-dive-into-yann-lecuns-jepa.html},
  identifies that JEPA's critic is ``trained from past states and subsequent
  intrinsic cost, retrieved from memory'' --- memory is load-bearing for the
  critic.

\bibitem{bjork1989forgetting}
Bjork, R.A.: Retrieval inhibition as an adaptive mechanism in human memory. In:
  Roediger, H.L., Craik, F.I.M. (eds.) Varieties of Memory and Consciousness:
  Essays in Honour of Endel Tulving, pp. 309--330. Lawrence Erlbaum Associates
  (1989)

\bibitem{burnell2026deepmind}
Burnell, R., Yamamori, Y., Firat, O., Olszewska, K., Hughes-Fitt, S., Kelly,
  O., Galatzer-Levy, I.R., Morris, M.R., Dafoe, A., Snyder, A.M., Goodman,
  N.D., Botvinick, M., Legg, S.: Measuring progress toward {AGI}: A cognitive
  framework. Tech. rep., Google DeepMind (March 2026),
  \url{https://storage.googleapis.com/deepmind-media/DeepMind.com/Blog/measuring-progress-toward-agi/measuring-progress-toward-agi-a-cognitive-framework.pdf},
  10-faculty cognitive taxonomy. \S7.5 Memory subdivides into
  Semantic/Episodic/Procedural/Prospective/Forgetting. Working Memory placed
  under \S7.8 Executive Functions, not Memory. Independent validation of the
  category error diagnosis.

\bibitem{cheng2026sycophancy}
Cheng, M., et~al.: Sycophantic {AI} decreases prosocial intentions and promotes
  dependence. Science  \textbf{391}(6792) (2026). \doi{10.1126/science.aec8352}

\bibitem{chhikara2025mem0}
Chhikara, P., et~al.: Mem0: Building production-ready {AI} agents with scalable
  long-term memory. arXiv preprint arXiv:2504.19413  (2025)

\bibitem{cohen1980procedural}
Cohen, N.J., Squire, L.R.: Preserved learning and retention of
  pattern-analyzing skill in amnesia: Dissociation of knowing how and knowing
  that. Science  \textbf{210}(4466),  207--210 (1980)

\bibitem{fronalabs2026frona}
{fronalabs}: Frona: Self-hosted personal {AI} assistant.
  \url{https://github.com/fronalabs/frona} (2026), rust + SurrealDB. Two-tier
  memory: user-scoped + agent-scoped. First Rust peer-competitor.

\bibitem{gulli2025agenticai}
Gulli, A., Sauco, M.: Agentic {AI}: Design Patterns and Production-Ready
  Strategies for Building Intelligent Agents. Springer (2025), 424 pages.
  Google Office of CTO. Two-tier memory (context + vector store). No temporal
  model, no forgetting, no wisdom layer, no consumer abstraction.

\bibitem{hawksight2026semantica}
{Hawksight-AI}: Semantica: Multi-backend bi-temporal knowledge engine.
  \url{https://github.com/Hawksight-AI/semantica} (2026), rDF/OWL + property
  graphs + triple stores. W3C PROV-O provenance. Allen Interval Algebra for
  deterministic temporal consistency.

\bibitem{hu2025memsurvey}
Hu, Y., Liu, S., Yue, Y., Zhang, G., et~al.: Memory in the age of {AI} agents:
  A survey --- forms, functions and dynamics  (2025)

\bibitem{jiang2026anatomymemory}
Jiang, D., Li, Y., Wei, S., Yang, J., Kishore, A., Zhao, A., Kang, D., Hu, X.,
  Chen, F., Li, Q., Li, B.: Anatomy of agentic memory: Taxonomy and empirical
  analysis of evaluation and system limitations. arXiv preprint
  arXiv:2602.19320  (2026)

\bibitem{karpathy2026llmkb}
Karpathy, A.: {LLM} knowledge bases. X (Twitter) (April 2026), widely shared
  (millions of views). Six-component system: ingest, compilation, Q\&A, output
  loop, linting. Independent convergence toward Knowledge layer with no Memory,
  Wisdom, or temporal semantics.

\bibitem{latimer2025hindsight}
Latimer, C., Boschi, N., Neeser, A., Bartholomew, C., Srivastava, G., Wang, X.,
  Ramakrishnan, N.: Hindsight is 20/20: Building agent memory that retains,
  recalls, and reflects (2025), bEAM SOTA: 64.1\% at 10M tokens. Five-level
  hierarchy. No forgetting, no bi-temporal, no supersession.

\bibitem{lecun2022jepa}
LeCun, Y.: A path towards autonomous machine intelligence. OpenReview  (2022),
  \url{https://openreview.net/pdf?id=BZ5a1r-kVsf}

\bibitem{lett2024jepa}
Lett, M.: Critical review of {LeCun's} introductory {JEPA} paper. Medium blog
  (2024),
  \url{https://medium.com/@malcolmlett/critical-review-of-lecuns-introductory-jepa-paper-f4e5e582caeb},
  identifies: configurator undefined, System 1/2 mapping incorrect, H-JEPA
  incompatible with predictive coding.

\bibitem{mastra2026om}
{Mastra}: Mastra observational memory. GitHub (2026),
  \url{https://github.com/mastra-ai/mastra}, compression-first memory: raw
  messages $\to$ observations $\to$ reflections. 94.87\% on LongMemEval
  (GPT-5-mini), 84.23\% with GPT-4o. Per-category: knowledge-update 96.2\%,
  temporal-reasoning 95.5\%, multi-session 87.2\% (ceiling across all systems).
  No cross-session persistence, no supersession.

\bibitem{mcdaniel2007prospective}
McDaniel, M.A., Einstein, G.O.: Prospective memory: An overview and synthesis
  of an emerging field. Sage Publications (2007)

\bibitem{minns2026minnsdb}
{Minns-ai}: {MinnsDB}: Agentic database with bi-temporal graph.
  \url{https://github.com/Minns-ai/MinnsDB} (2026), aGPL-3.0, Rust. Bi-temporal
  knowledge graph. MinnsQL with \texttt{WHEN}/\texttt{AS OF}. OWL vocabulary as
  metadata, not reasoning.

\bibitem{morris2026jaggedness}
Morris, M.R., Altman, D., Belfield, H., Goemans, A., Iqbal, H., Burnell, R.,
  Gabriel, I., Albanie, S., Dafoe, A.: Characterizing model jaggedness supports
  safety and usability. Tech. rep., Google DeepMind (January 2026), monolithic
  scores hide jagged capability profiles; principled decomposition required.

\bibitem{nelson1990metamemory}
Nelson, T.O.: Metamemory: A theoretical framework and new findings. In: Bower,
  G.H. (ed.) The Psychology of Learning and Motivation, vol.~26, pp. 125--173.
  Academic Press (1990)

\bibitem{signet2026}
{niloproject}: Signet: Persistent cognition layer for {AI} coding agents.
  GitHub (2026), entity/aspect/attribute graph with partial supersession but
  uniform 0.95\^{}days decay on all types.

\bibitem{nornicdb2026}
{NornicDB}: {NornicDB}: Cognitive graph and vector database for {AI} agents.
  GitHub (2026), three-tier cognitive decay: episodic 7-day, semantic 69-day,
  procedural 693-day half-life. Applies storage-level decay to factual
  knowledge.

\bibitem{nous2026hermesagent}
{Nous Research}: Hermes agent.
  \url{https://github.com/NousResearch/hermes-agent} (2026), five memory layers
  including Procedural Skill Documents following the \texttt{agentskills.io}
  open standard. Closes experience-to-wisdom loop.

\bibitem{oriminemos2026}
{ori-community}: Ori mnemos v0.4: Six-layer retrieval with graph-aware
  forgetting. GitHub (2026), three-zone decay (identity 0.1x, knowledge 1.0x,
  ops 3.0x). ACT-R vitality, Tarjan structural protection. Still conflates all
  data types in one graph.

\bibitem{packer2023memgpt}
Packer, C., Wooders, S., Lin, K., Fang, V., Patil, S.G., Stoica, I., Gonzalez,
  J.E.: {MemGPT}: Towards {LLMs} as operating systems  (2023), virtual context
  management for LLMs via OS-inspired memory hierarchy. Letta V1 rearchitecture
  (2025) introduces Context Repositories and tiered memory.

\bibitem{papr2026sdk}
{Papr-ai}: Papr python {SDK}: Schema-policy memory engine.
  \url{https://github.com/Papr-ai/papr-pythonSDK} (2026), schema-policy DSL
  decorators for memory operations. First-class LLM callouts at
  constraint-firing time.

\bibitem{mempalace2026drama}
{r/AIMemory community}: Proposal: A real benchmark for long-term {AI} memory
  systems. \url{https://www.reddit.com/r/AIMemory/comments/1sgvsxb/} (April
  2026), memPalace audit: reported 96.6\% LongMemEval collapsed to 66.8\%;
  LoCoMo answer key 6.4\% wrong; LLM judge accepts 63\% of incorrect answers.

\bibitem{rasmussen2025graphiti}
Rasmussen, P., Paliychuk, P., Beauvais, T.: Zep: A temporal knowledge graph
  architecture for agent memory  (2025), bi-temporal model (4 timestamps per
  edge). Episode-to-fact extraction.

\bibitem{rohitg002026llmwiki}
{rohitg00}: {LLM} wiki v2.
  \url{https://gist.github.com/rohitg00/2067ab416f7bbe447c1977edaaa681e2}
  (2026), fork of Karpathy pattern. Confidence scoring with time-decay,
  supersession, Ebbinghaus forgetting, four-tier consolidation. Closest
  informal statement of the four-layer thesis.

\bibitem{google2026aoma}
Saboo, S.: Always-on memory agent. GitHub (GoogleCloudPlatform/generative-ai)
  (2026),
  \url{https://github.com/GoogleCloudPlatform/generative-ai/tree/main/gemini/agents/always-on-memory-agent},
  mIT. No vector DB. LLM-driven memory organization in SQLite. 30-minute
  consolidation loops generating meta-insights. Independently validates the
  consolidation-as-first-class-operation pattern.

\bibitem{sumers2024coala}
Sumers, T.R., Yao, S., Narasimhan, K., Griffiths, T.L.: Cognitive architectures
  for language agents. Transactions on Machine Learning Research (TMLR)
  (2024), \url{https://arxiv.org/abs/2309.02427}

\bibitem{tavakoli2026beam}
Tavakoli, M., Salemi, A., Ye, C., Abdalla, M., Zamani, H., Mitchell, J.R.:
  {BEAM}: Beyond a million tokens: Benchmarking and enhancing long-term memory
  in {LLMs}. arXiv preprint arXiv:2510.27246  (2026)

\bibitem{tulving1972episodic}
Tulving, E.: Episodic and semantic memory. In: Tulving, E., Donaldson, W.
  (eds.) Organization of Memory, pp. 381--403. Academic Press (1972)

\bibitem{yu2026agemem}
Yu, Y., Yao, L., Xie, Y., Tan, Q., Feng, J., Li, Y., Wu, L.: Agentic memory:
  Learning unified long-term and short-term memory management for large
  language model agents  (2026), ebbinghaus forgetting curve adaptation for AI
  memory.

\bibitem{zhang2025amem}
Zhang, W., et~al.: A-mem: Agentic memory for {LLM} agents  (2025)

\end{thebibliography}

\end{document}